\pdfoutput=1
\documentclass[conference]{IEEEtran}
\IEEEoverridecommandlockouts
\usepackage{cite}
\usepackage{amsmath,amssymb,amsfonts}
\usepackage{algorithmic}
\usepackage{graphicx}
\usepackage{textcomp}
\usepackage{xcolor}
\usepackage{multirow}
\usepackage{times}
\usepackage{hyperref}
\usepackage[numbers,sort&compress]{natbib}
\def\BibTeX{{\rm B\kern-.05em{\sc i\kern-.025em b}\kern-.08em
    T\kern-.1667em\lower.7ex\hbox{E}\kern-.125emX}}

\makeatletter
\renewcommand{\maketag@@@}[1]{\hbox{\m@th\normalsize\normalfont#1}}%
\makeatother

\begin{document}

\title{Parallel Physics-Informed Neural Networks with Bidirectional Balance}

\author{\IEEEauthorblockN{Yuhao Huang\IEEEauthorrefmark{1}}
\vspace{0.2cm}
\IEEEauthorblockA{\IEEEauthorrefmark{1}Beijing Jiaotong University, Beijing, China\\
21121552@bjtu.edu.cn}
}

\maketitle

\begin{abstract}
    As an emerging technology in deep learning, physics-informed neural networks (PINNs) have been widely used to solve various partial differential equations (PDEs) in engineering. However, PDEs based on practical considerations contain multiple physical quantities and complex initial boundary conditions, thus PINNs often returns incorrect results. Here we take heat transfer problem in multilayer fabrics as a typical example. It is coupled by multiple temperature fields with strong correlation, and the values of variables are extremely unbalanced among different dimensions. We clarify the potential difficulties of solving such problems by classic PINNs, and propose a parallel physics-informed neural networks with bidirectional balance. In detail, our parallel solving framework synchronously fits coupled equations through several multilayer perceptions. Moreover, we design two modules to balance forward process of data and back-propagation process of loss gradient. This bidirectional balance not only enables the whole network to converge stably, but also helps to fully learn various physical conditions in PDEs. We provide a series of ablation experiments to verify the effectiveness of the proposed methods. The results show that our approach makes the PINNs unsolvable problem solvable, and achieves excellent solving accuracy.
\end{abstract}


\section{Introduction}
With the rapid development of universal data and computing resources, deep learning technology has produced abundant achievements 
in different subjects \cite{2012ImageNet,2014Learning,2015Predicting}. Due to the universal approximation capability of neural 
networks, related technologies have also been affecting the fields of computing science and engineering. As early as the 1990s, 
artificial neural network learned convective heat transfer coefficient from data \cite{1996Evaluating}. Further, Owhadi 
\cite{owhadi2015bayesian} exploited prior knowledge to make an attempt in numerical homogenization problem. Convolutional neural 
network was also used for explore heat transport properties of turbulent Rayleigh-Benard convection \cite{fonda2019deep}. Obviously, 
these supervised learning tasks cannot be used for regular numerical solving, because the ground truth are often unknown. In 
subsequent studies, Raissi et al. \cite{raissi2019physics} proposed physics-informed neural networks (PINNs), in which the losses 
are derived from physical conditions. This pioneering work has achieved plenty of remarkable results, including hydrodynamics 
\cite{raissi2020hidden,sun2020surrogate,raissi2019deep} bioengineering \cite{costabal2020physics,kissas2020machine}, high-dimensional 
PDEs \cite{sirignano2018dgm,han2018solving} and etc. 
\par
However, PINNs usually cannot be correctly solved in practical engineering problems, especially when the data distribution is highly 
uneven \cite{sun2020surrogate,fuks2020limitations}. Wang et al. \cite{wang2020understanding} believed that gradient pathology will 
occurr when data showed high-frequency. They proposed a self-adaptive method which adjust weights of different terms in composite 
loss function. Although this method improves accuracy significantly, it cannot generalize to other possible data situation. Just 
like the common heat transfer of clothing in daily life \cite{udayraj2016heat}, it is accompanied by data imbalance between 
multiple dimensions. PINNs will have abnormal training feedback with extremely high losses and unrealistic results in this situation, 
and there is no relevant research. Hence, we take this problem in a more extreme environment as a prime example to tap more 
potential of PINNs.
\par
Compared with the development of PINNs, the study of numerical heat transfer is obviously much more mature and complete. At first, 
Torvi and Dale \cite{torvi1999heat} clearly summarized heat transfer process in thin fabric under high heat flux. Chitrphiromsri 
and Kuznetsov \cite{chitrphiromsri2003modeling} analyzed simultaneous heat and moisture transfer through fabric. Further, Zhu and 
Zhang \cite{zhu2009modeling} added pyrolysis of fabrics and related shrinkage on the basis of Torvi's work. More comprehensively, 
Ghazy and Bergstrom \cite{ghazy2012numerical}  fully considered the heat transfer in air gap. As we can see, when the related 
mathematical models are getting closer and closer to reality, traditional numerical methods are also getting bloated. On the 
contrary, PINNs is usually not affected by the complexity of equations due to its special mechanism, that it transforms the 
iterative solving into the parameters learning. It shows great potential in numerical heat transfer. In this way, it is very 
valuable to solve the training pathology in PINNs.
\par
Based on the above discussion, we conduct research on the application of PINNs to thermal protective clothing. Our specific 
contributions can be summarized in the following points:
\begin{itemize}
\item Our analysis points out that the failure of PINNs is related to the numerical imbalance in data forward and gradient back-propagation processes.
\item We solve the coupled PDEs system in parallel by combining multiple neural sub-networks, which has stronger fitting ability.
\item We propose the Forward Balance Module which maintain the forward value of the overall network units within a reasonable range. Correspondingly, the Backward Balance Module is proposed to pay balanced attention to various physical constraints by scaling losses.
\item Our approach can solve the thermal protective clothing problem with high performance, making the PINNs unsolvable problem solvable.
\end{itemize}
\par
Taken together, our development provides a new perspective for the training of constrained neural networks, which can help us endow deep learning tools with prior knowledge and reduce barriers in extending to other scientific fields.

\section{Preliminaries}
In this section, we will briefly review the classic PINNs and the mathematical model of thermal protective clothing. These preliminaries will help the subsequent theoretical analysis and method proposed. 
\begin{figure}[htbp]
  \centering
  \includegraphics[width=\linewidth]{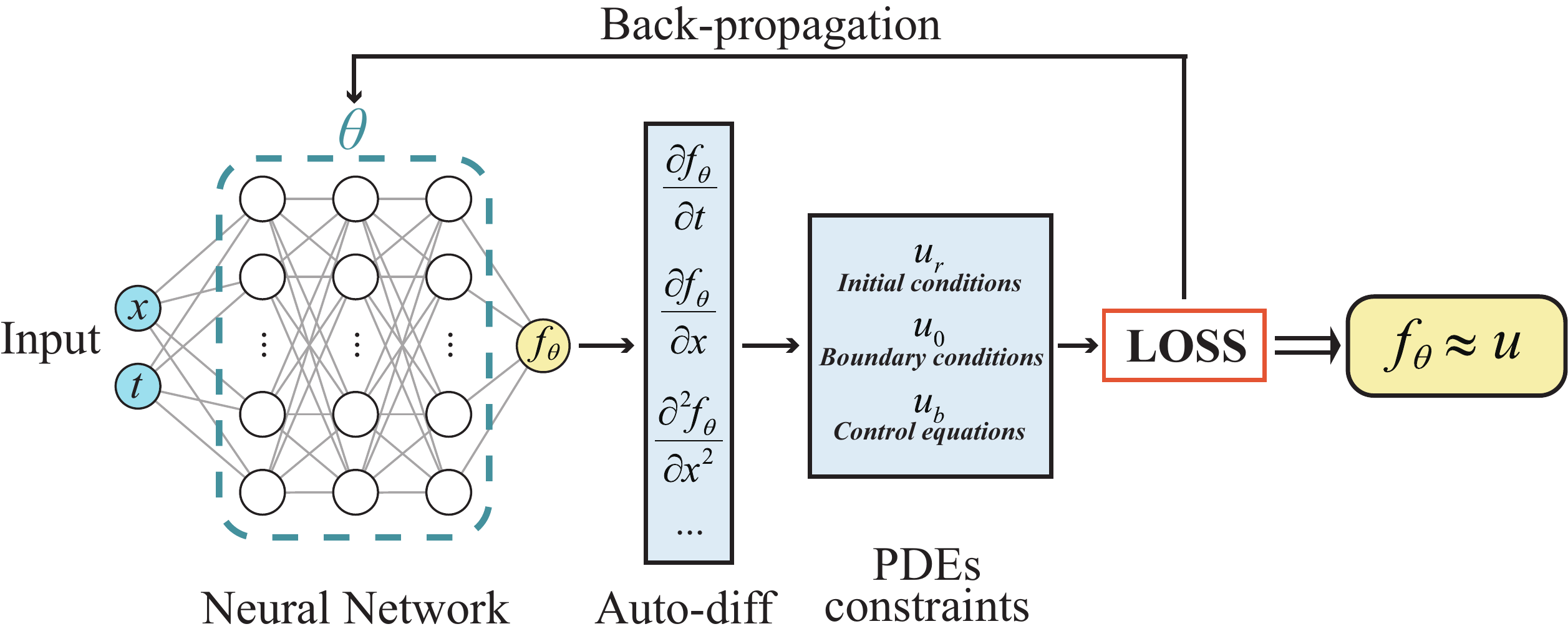}
  \caption{Overview of the physics-informed neural networks (PINNs).}
  \label{fig1}
\end{figure}
\subsection{Physics-informed Neural Networks}
PINNs is designed to solve forward and inverse problems of PDEs. Generally speaking, the mixed initial-boundary value problem can be summarized as:
\begin{subequations}
\begin{align}
  &\boldsymbol{u}_{t}+\mathcal{N}_{x}[\boldsymbol{u}]=0, \quad x \in \Omega, \quad t \in[0, T] \label{1a}\\
  &\boldsymbol{u}(x, 0)=h(x), \quad x \in \Omega \label{1b}\\
  &\boldsymbol{u}(x, t)=g(x, t), \quad t \in[0, T], \quad x \in \partial \Omega \label{1c}
\end{align}
\end{subequations}
where, $(x,t)$ is the coordinates in the finite computational domain; $\boldsymbol{u}$ is the solution of 
this PDE; $\mathcal{N}_{x}$ is a general linear or nonlinear differential operator in control equation; 
The initial and boundary condition can be expressed as in \eqref{1b}-\eqref{1c}. This typical PDEs covers a 
range of problems in mathematical physics, including conservation laws, diffusion processes and 
advection-diffusion-reaction systems.
\par
According to the original work of PINNs \cite{raissi2019physics}, we can fit analytical solution $\boldsymbol{u}(x, t)$ by neural 
network with parameters $\theta$ (weight or bias). The network takes discrete points $(x,t)$ as inputs and 
corresponding value $f_{\theta}(x,t)$ as outputs. The learning of parameters is constrained by the physical 
properties in PDEs (control equations, initial conditions, boundary conditions), which can be defined as 
the following residual.
\begin{subequations}
\begin{align}
  &u_{r}(x, t):=\frac{\partial}{\partial t} f_{\theta}(x, t)+\mathcal{N}_{x}\left[f_{\theta}(x, t)\right] \label{2a}\\
  &u_{o}(x, 0):=f_{\theta}(x, 0)-h(x) \label{2b}\\
  &u_{b}(x, t):=f_{\theta}(x, t)-g(x, t) \label{2c}
\end{align}
\end{subequations}
\par
The partial derivative term can be quickly calculated by automatic differential technique \cite{paszke2019pytorch}. 
Thus, $\theta$ can be trained by minimizing a composite loss function as follows.
\begin{subequations}
\begin{align}
  &\mathcal{L}(\theta):=\sum_{i} \mathcal{L}_{i}(\theta) \label{3a}\\
  &\mathcal{L}_{r}=\frac{1}{N_{r}} \sum_{j=1}^{N_{r}}\left[u_{r}\left(x_{r}^{j}, t_{r}^{j}\right)\right]^{2} \label{3b}\\
  &\mathcal{L}_{0}=\frac{1}{N_{0}} \sum_{j=1}^{N_{0}}\left[u_{o}\left(x_{0}^{j}, 0\right)\right]^{2} \label{3c}\\
  &\mathcal{L}_{b}=\frac{1}{N_{b}} \sum_{j=1}^{N_{b}}\left[u_{b}\left(x_{b}^{j}, t_{b}^{j}\right)\right]^{2} \label{3d}
\end{align}
\end{subequations}
Here, $(x_0^j,0)$ denotes the data under the initial conditions, $(x_b^j,t_b^j)$ denotes the data under the 
boundary conditions, and $(x_r^j,t_r^j)$ denotes the remaining data in the entire computational domain. 
All loss terms used mean square error loss as in \eqref{3b}-\eqref{3d}.
\par
Based on these, PINNs uses soft constraints to make predictions that satisfy any conditions derived from physical law such as symmetry, invariance and conservation.
\begin{figure}[htbp]
  \centering  
  \includegraphics[width=\linewidth]{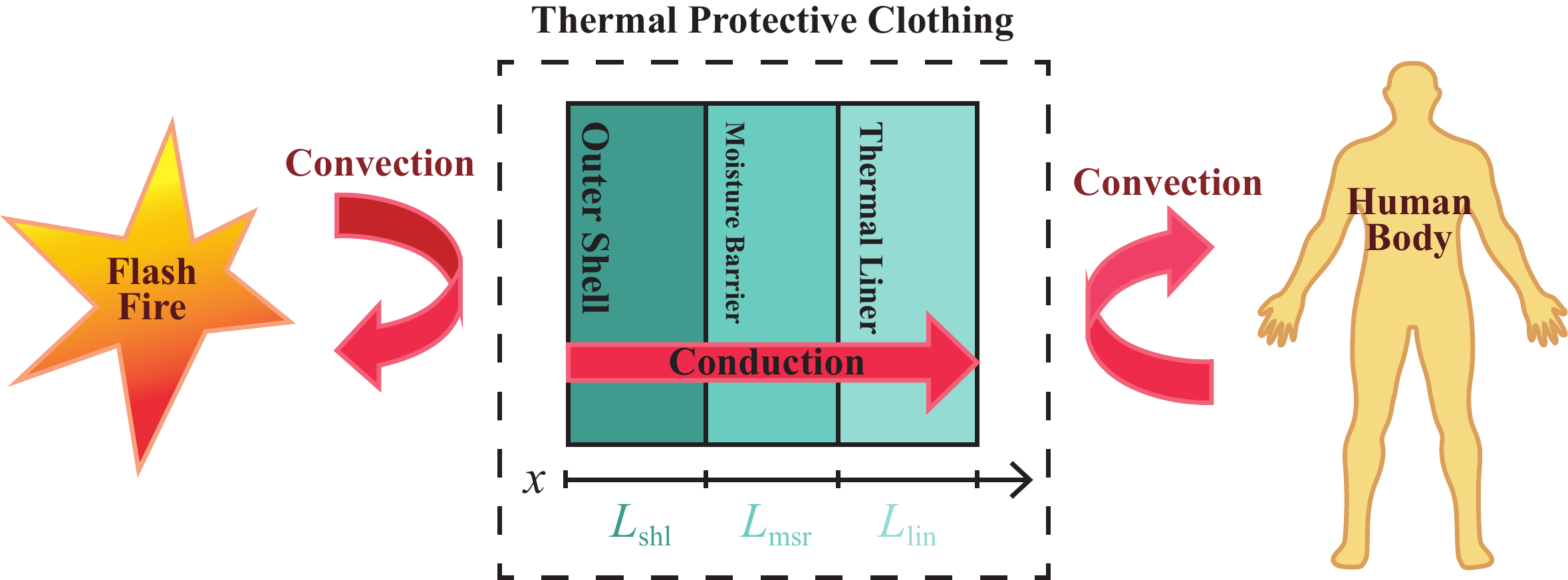}  
  \caption{Schematic of the heat transfer process in thermal protective clothing}
  \label{heattransfer}
\end{figure}

\subsection{Heat Transfer Model of Multi-layer Fabric}
\label{sec:htm}
\begin{figure*}[t]
  \centering  
  \includegraphics[width=\textwidth]{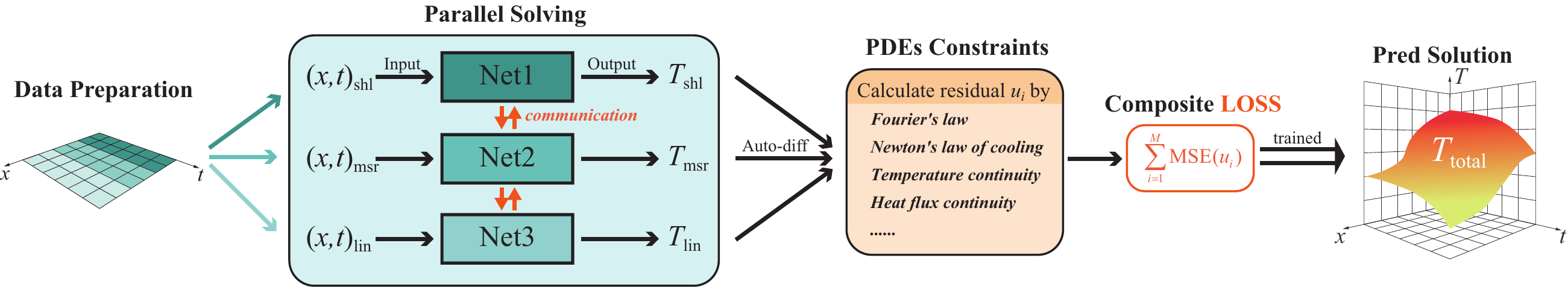}  
  \caption{Overview of parallel solution framework (PSF). In the data preparation stage, the input grid will be divided into several sub-domains. These datasets will be input into the parallel network to generate the corresponding solutions, and the PDEs constraints will be obtained by the automatic differentiation technique (auto-diff), and then the total loss function will be calculated. Through iterative training, a global prediction solution with the expected accuracy is obtained.}
  \label{PSF}
\end{figure*}

A typical thermal protective clothing \cite{udayraj2016heat} is a multilayer arrangement (usually 3 layers: outer shell, moisture 
barrier and thermal liner) as shown in Fig.~\ref{heattransfer}. The outer surface of the shell is exposed to flash fire, and energy 
is transmitted to the shell by thermal convection. Further, heat passes through other fabric layers by conduction. In order to 
clearly introduce the focus of this manuscript, we simplify the heat transfer problem to one-dimensional and assume that the 
materials of each layer are isotropic. According to Fourier's Law, we can build control equations for thermal diffusion reaction 
as follow \cite{torvi1999heat}.
\begin{align}
  C_{\text{shl}}^{\text{A}} \frac{\partial T_{\text{shl}}}{\partial t}&=k_{\text{shl}} \frac{\partial^{2} T_{\text{shl}}}{\partial x^{2}}\label{4}\\
  C_{\text{msr}}^{\text{A}} \frac{\partial T_{\text{msr}}}{\partial t}&=k_{\text{msr}} \frac{\partial^{2} T_{\text{msr}}}{\partial x^{2}}\label{5}\\
  C_{\text{lin}}^{\text{A}} \frac{\partial T_{\text{lin}}}{\partial t}&=k_{\text{lin}} \frac{\partial^{2} T_{\text{lin}}}{\partial x^{2}}\label{6}
\end{align}
where, the subscripts shl, msr and lin indicate the outer shell, the moisture barrier and the thermal liner, 
respectively; $T$ is the temperature value $(\text{K})$; $x$ and $t$ are spatial $(\text{m})$ and temporal 
$(\text{s})$ coordinates, respectively; $C^{\text{A}}$ is the apparent heat capacity $(\text{J} / (\text{m}^3 \cdot \text{K}))$; 
$k$ is the thermal conductivity $(\text{W} / (\text{m} \cdot \text{K}))$.
\par
Assuming that $T_0$ is the normal temperature, the coupled system meets the same initial condition as follow.
\begin{equation}
  T_{\text{shl}}|_{t=0} = T_{\text{msr}}|_{t=0} = T_{\text{lin}}|_{t=0} = T_0 \label{7}
\end{equation}

\par
For the left and right boundary of the whole system, i.e. the outer surface of the outer shell and the inner 
surface of the thermal liner, there is convective heat transfer from air. The difference is that the outer 
shell is in contact with the hot gases $T_{\text{g}}$ in flash fire environment, while the thermal liner is in contact 
with the air $T_0$ in ambient. According to Newton's law of cooling, the Neumann boundary conditions are 
given as follows:
\begin{align}
  -\left.k_{\text{shl}} \frac{\partial T_{\text{shl}}}{\partial x}\right|_{x=0}&=h_{\text{g}}\left(T_{\text{g}}-\left.T_{\text {shl}}\right|_{x=0}\right) \label{8}\\
  -\left.k_{\text{lin}} \frac{\partial T_{\text{lin}}}{\partial x}\right|_{x=L_{\text{fab}}}&=h_{\text{air}}\left(\left.T_{\text{lin}}\right|_{x=L_{\text{fab}}}-T_{0}\right) \label{9}
\end{align}
where, $L_{\text{fab}}$ is the total thickness of all fabric layers; $h_{\text{g}}$ and $h_{\text{air}}$ are the convective heat transfer coefficients from flame to the 
outer shell and from ambient to the thermal liner, respectively.
\par
Since the boundary between adjacent layers satisfies the same temperature and heat flux, we can obtain the 
following Dirichlet and Neumann boundary conditions for adjacent layers \cite{ghazy2012numerical}.
\begin{align}
  \left.T_{\text{shl}}\right|_{x=L_{\text{shl}}}&=\left.T_{\text{msr}}\right|_{x=L_{\text{shl}}} \label{10}\\
  -\left.k_{\text{shl}} \frac{\partial T_{\text{shl}}}{\partial x}\right|_{x=L_{\text{shl}}}&=-\left.k_{\text{msr}} \frac{\partial T_{\text{msr}}}{\partial x}\right|_{x=L_{\text{shl}}} \label{11}\\
  \left.T_{\text{msr}}\right|_{x=L_{\text{msr}}}&=\left.T_{\text{lin}}\right|_{x=L_{\text{msr}}} \label{12}\\
  -\left.k_{\text{msr}} \frac{\partial T_{\text{msr}}}{\partial x}\right|_{x=L_{\text{msr}}}&=-\left.k_{\text{lin}} \frac{\partial T_{\text{lin}}}{\partial x}\right|_{x=L_{\text{msr}}} \label{13}
\end{align}
\par
In summary, the heat transfer model in multilayer fabric is established by such fixed solution problem as in \eqref{4}-\eqref{13}.

\section{Methodology}
The failure of PINNs is often accompanied by abnormal loss values and unrealistic prediction results. In this section, we deeply analyze the causes of this failure and why our methods can work successfully.
\subsection{Parallel Solving Framework}
\label{sec:parallel solving framework}
As mentioned earlier, heat transfer model in multilayer fabric is actually the coupled PDEs problem. Each 
fabric layer cannot be viewed as an independent heat transfer system due to the conditions that the 
temperature and heat flux of adjacent boundary are consistent. On the other hand, we can't construct a 
well-posed problem without these boundary conditions \cite{1995THE}.
\par
In order to solve this kind of coupled PDEs problem, we propose the Parallel Solving Framework (PSF) of PINNs as shown in Fig.~\ref{PSF}, 
which fits the temperature function of each layer in parallel to make the interdependent boundary conditions update dynamically 
in the training process. For three fabrics, we give three MLP to fit the temperature field respectively. These sub-networks use 
the same structure, but their parameters are not shared. Similar to \eqref{3a}, 
the residual terms of all equations in the coupled PDEs are involved in the composite loss function as follow. 
where $\mathcal{L}(\cdot)$ is mean square error loss function. You can find the details for this part in Appendix \ref{app:res}.

\vspace{-0.2cm}
\begin{small}
\begin{align}
  \text{LOSS}=&\sum \mathcal{L}\left(u_{i}^{\text{layer}}\right) \notag\\
  =&\mathcal{L}\left(u_{r}^{\text{shl}}\right)+\mathcal{L}\left(u_{0}^{\text {shl}}\right)+
  \mathcal{L}\left(u_{b}^{\text {shl}}\right)+\mathcal{L}\left(u_{b1}^{\text {shl\&msr}}\right)+ \notag\\
  &\mathcal{L}\left(u_{b2}^{\text {shl\&msr}}\right)+\mathcal{L}\left(u_{r}^{\text{msr}}\right)+
  \mathcal{L}\left(u_{0}^{\text{msr}}\right)+\mathcal{L}\left(u_{b1}^{\text {msr\&lin}}\right)+ \notag\\
  &\mathcal{L}\left(u_{b2}^{\text {msr\&lin}}\right)+\mathcal{L}\left(u_{r}^{\text{lin}}\right)+
  \mathcal{L}\left(u_{0}^{\text{lin}}\right)+\mathcal{L}\left(u_{b}^{\text{lin}}\right) \label{14}
\end{align}
\end{small}

\par
According to different loss terms, the calculating domain is divided into multiple sub-domains. Then, we use the gradient descent 
algorithm to optimize the parameters of all sub-networks simultaneously. It is worth noting that this PINNs framework is a global 
unsupervised training, which only relies on existing constraints derived from physical laws to learn.
\par
The loss terms of adjacent layer, such as $\mathcal{L}\left(u_{b1}^{\text {shl\&msr}}\right)$, 
are slightly different from other terms, because their gradients back-propagation will inevitably affect two adjacent sub-networks, 
which allows communication between sub-networks. Such a mechanism provides a fault-tolerant space for the initial stage of 
unsupervised training, and further enables all sub-networks to adjust and learn together.

\subsection{Forward Balance Module}
\label{sec:forward balance module}
\begin{figure*}[htbp]
    \centering  
    \includegraphics[width=0.9\textwidth]{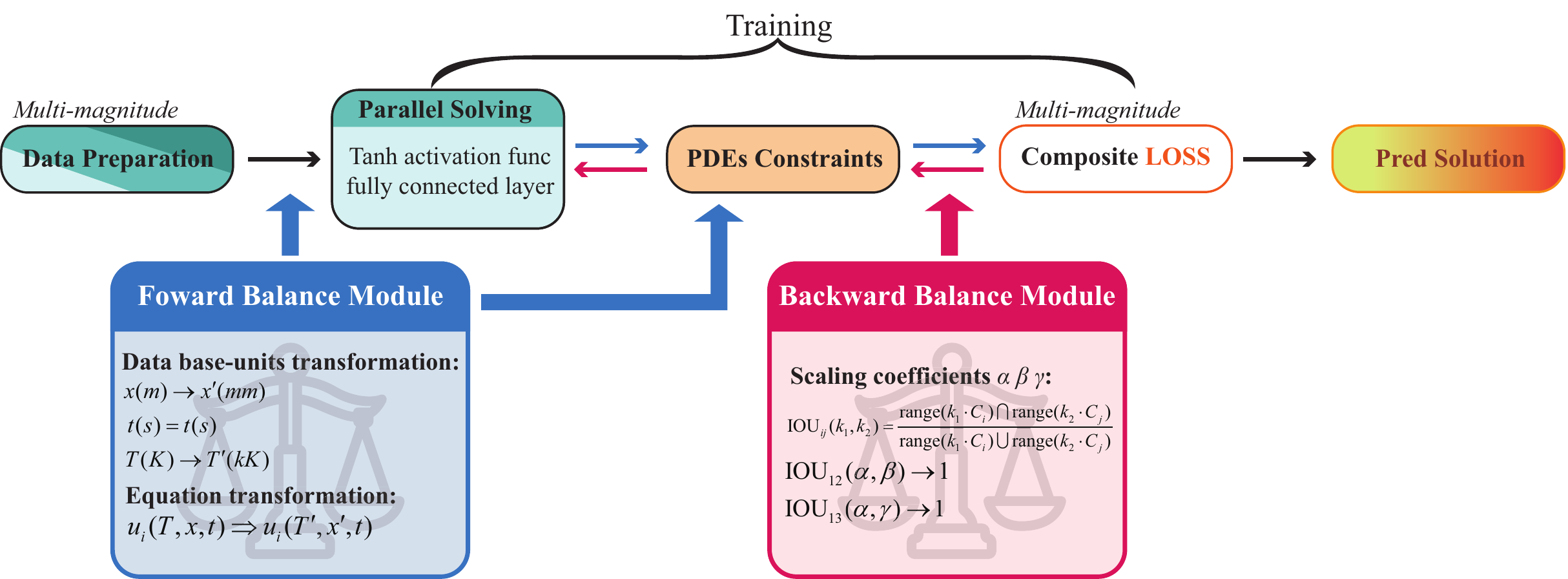}  
    \caption{Schematic of the bidrectional balance method. Blue and red respectively correspond to the forward and backward processes.}
    \label{balance}
\end{figure*}
As the input of PINNs, the fabric thickness is usually a few tenths of a millimeter, and the flash time can reach tens of seconds. 
But as the output, the temperature can reach several hundred Kelvin theoretically. Such numerical imbalance brings higher 
requirements for neural networks in large-scale mapping and nonlinear fitting.
\par
Despite the existence of the universal approximation theorem \cite{hornik1989multilayer}, not any network structure can meet our 
needs. The activation function profoundly affects the performance of the shallow network \cite{siegel2020approximation}. Among 
them, Tanh-like functions restrict the output to a small range, while ReLU-like functions restrict the gradient through linear 
units to alleviate gradient disappearance. Obviously, the former is not conducive to large-scale mapping, and the latter has poor 
nonlinear fitting ability in shallow networks.
\par
Based on this, the Forward Balance Module (FBM) scales the data pairs losslessly to a similar range, so that the shallow network 
with Tanh that has sufficient nonlinear fitting capabilities does not need to focus on large-scale mapping. Specifically, the FBM 
first transforms the physical base-units of input in PDEs. The unit of length $x$ is converted from meter $(\text{m})$ to 
millimeter $(\text{mm})$, and the unit of time $t$ remains unchanged. Then, we use equation (soft constraint) transformation to 
indirectly turn the unit of the output temperature value $T$ from Kelvin $(\text{K})$ to kilokelvin $(\text{kK})$. Taking the 
outer shell as an example, the change of the residual of the control equation is given as follow.
\begin{align}
  u_{r}^{\text{shl}}
  &= C_{\text{shl}}^{\text{A}} \frac{\partial T_{\text{shl}}}{\partial t}-k_{\text{shl}}\frac{\partial^{2} T_{\text{shl}}}{\partial x^{2}} \notag\\
  &= C_{\text{shl}}^{\text{A}} \frac{\partial \left(T'_{\text{shl}} \times 10^{3} \right)}{\partial t}-
  k_{\text{shl}}\frac{\partial^{2} \left(T'_{\text{shl}}\times 10^{3} \right)}{\partial {\left(x' \times 10^{-3} \right)}^{2}} \notag\\
  &= C_{\text{shl}}^{\text{A}} \frac{\partial T'_{\text{shl}}}{\partial t} \times 10^{3}-k_{\text{shl}} \frac{\partial^{2} T'_{\text{shl}}}{\partial x'^2} \times 10^{9} \label{15}
\end{align}
\par
FBM is placed in the data preparation stage of the overall solving framework (see Fig.~\ref{balance}). It balances the values in 
the subsequent neural network just like normalization or nondimensionalization. The difference is that FBM maintains equivalence 
compared to normalization, and can still retain physical meaning compared to nondimensionalization.

\subsection{Backward Balance Module}
\label{sec:backward balance module}
The PINNs-based solving approach relies on soft constraints derived from physical laws. Such constraints directly affect the 
gradient flow, making the network parameters update in the direction of gradient descent. This process can be described as the 
following forward Euler discretization.
\begin{align}
  \theta_{n+1} &=\theta_{n}-\eta \nabla_{\theta_{n}} \text {LOSS} \notag\\
  &=\theta_{n}-\eta \sum \nabla_{\theta_{n}} \mathcal{L}\left(u_{i}^{\text{layer}}\right) \label{16}
\end{align}
where, $\eta$ is the learning rate, and the gradient $\nabla_{\theta_{n}} \mathcal{L}\left(u_{i}^{\text{layer}}\right)$ 
usually increases as the loss term $\mathcal{L}\left(u_{i}^{\text{layer}}\right)$ increases. When the residual 
terms $u_{i}^{\text{layer}}$ is unbalanced during training, the gradient flow will be affected by this, 
causing the neural network to pay more attention to larger terms \cite{wang2020understanding}. 
It is well known that PDEs may have infinitely many solutions without proper initial-boundary conditions \cite{evans2010partial}, 
so PINNs with learning imbalance often returns incorrect predictions.

\par
Back to the heat transfer problem to explore the imbalance in the backward process of the network. By 
comparing expressions in Appendix \ref{app:res}, we can find that so many residual terms can also be summarized 
as following partition. where, the elements of $C_1$ all contain $T$ term, $C_2$ all contain $\frac{\partial T}{\partial t}$ 
term, and $C_3$ all contain $\frac{\partial^{2} T}{\partial x^{2}}$ term. 
\begin{subequations}
\begin{align}
  &C_{1}=\left\{u_{0}^{\text{shl}}, u_{0}^{\text{msr}}, u_{0}^{\text{lin}}, u_{b1}^{\text{shl\&msr}}, u_{b1}^{\text{msr\&lin}}\right\} \label{17a}\\
  &C_{2}=\left\{u_{b}^{\text{shl}}, u_{b2}^{\text{shl\&msr}}, u_{b2}^{\text{msr\&lin}}, u_{b}^{\text{lin}}\right\} \label{17b}\\
  &C_{3}=\left\{u_{r}^{\text{shl}}, u_{r}^{\text{msr}}, u_{r}^{\text{lin}}\right\} \label{17c}
\end{align}
\end{subequations}

\par
Such a rule indicates the existence of learning imbalance. To quantify this imbalance, We define the range of $C_i$ as 
$\text{range}\left(C_{i}\right)=\left[\min \left(\mathcal{L}\left(C_{i}\right)\right), \max \left(\mathcal{L}\left(C_{i}\right)\right)\right]$, 
where $\mathcal{L}(C)=\left\{(\sum_{k=1}^{n_{\text{exp}}} \mathcal{L}\left(u_{j}\right)_{k})/n_\text{exp} \mid u_{j} \in C\right\}$ 
(i.e, a set composed of the mean square error of each element in $C$ under $n_\text{exp}$ repeated experiments.). If above three 
classes are balanced, their their $\text{range}\left(C_{i}\right)$ should also highly overlap. We define this degree of overlap 
with the help of Intersection of Union (IOU) as in \eqref{18}. When the IOU is closer to 1, the degree of overlap between sets 
is higher, and vice versa.

\vspace{-0.2cm}
\begin{small}
\begin{equation}
  \text{IOU}_{i j}\left(k_{1}, k_{2}\right)=\frac{\text{range}\left(k_{1} \cdot C_{i}\right) \cap \text{range}\left(k_{2} \cdot C_{j}\right)}
  {\text{range}\left(k_{1} \cdot C_{i}\right) \cup \text{range}\left(k_{2} \cdot C_{j}\right)} \label{18}
\end{equation}
\end{small}

\par
Based on these definitions, we propose the Backward Balance Module (BBM), which multiplies the three types of 
sets with scaling coefficients $\alpha, \beta, \gamma$ to make $\text{IOU}_{12}(\alpha, \beta) \rightarrow 1$ 
and $\text{IOU}_{13}(\alpha, \gamma) \rightarrow 1$. As shown in Fig.~\ref{balance}, the BBM is usually placed 
between the residual and the loss calculation. It balances the gradient flow in the back-propagation through 
the scaling operation, so that the network can fully learn the different physical laws in PDEs.

\section{Numerical Results}
In this section, we provide a series of experiments aimed at evaluating the solving performance in different PINNs structures. 
We develop a benchmark in all situations:
\par
First, we assume thermophysical properties and initial conditions of each fabric layer are shown in Tables \ref{Tab1} and \ref{Tab2}. 
Secondly, we divide the three fabric layers into 50, 70, and 200 segments respectively through a uniform grid, and divide whole 
time domain into 300 segments. This setting can generate a training dataset (network input coordinates) with size of (300×320). 
Then, the backbone is an MLP with 4 hidden layers with 10 channels. Lastly, this benchmark uses the default Adam optimizer \cite{kingma2015adam} 
(learning rate is 0.001) and the Kaiming random initialization method \cite{he2015delving}.

\begin{table}[htbp]
  \caption{Thermophysical properties of fabric}
  \label{Tab1}
  \vspace{-0.5cm}
  \begin{center}
      \resizebox{\linewidth}{!}{
      \begin{tabular}{|c|c|c|c|c|}
          \hline     
          \multirow{2}{*}{\textbf{Material}} & \textbf{Density} & \textbf{Specific heat} & \textbf{Heat conductivity} & \textbf{Thickness} \\        
          & $(\text{kg} / \text{m}^{3})$ & $(\text{J} / (\text{kg}\cdot \text{K}))$ & $(\text{W} / (\text{m}\cdot \text{K}))$ & $(\text{mm})$ \\
          \hline
          Outer shell & 300 & 1377 & 0.082 & 0.6 \\ 
          Moisture barrier & 862 & 2100 & 0.37 & 0.85\\
          Thermal liner & 74.2 & 1726 & 0.045 & 3.6 \\
          \hline
      \end{tabular}
      }
  \end{center}
\end{table}
\vspace{-0.5cm}
\begin{table}[htbp]
  \caption{Initial conditions for fabric and environment}
  \label{Tab2}
  \vspace{-0.5cm}
  \begin{center}
      \resizebox{\linewidth}{!}{
      \begin{tabular}{|c|c|}
      \hline
      \textbf{Property} & \textbf{Value} \\
      \hline
      {Normal temperature} & {310.15 K} \\
      {Hot gases temperature} & {2000 K} \\
      {Flame convective heat transfer coefficient} & {40 $\text{W} / (\text{m}^2 \text{K})$} \\
      {Air convective heat transfer coefficient} & {9.496 $\text{W} / (\text{m}^2 \text{K})$} \\
      \hline
      \end{tabular}
      }
  \end{center}
\end{table}

\par
We take the solutions of traditional finite difference method (FDM) as the ground truth. According to the stability condition 
based on the Fourier number, FDM generates a grid size of ($200000\times 320$), and solves PDEs by iterative calculation.

\par
All algorithms are mainly implemented by Pytorch. The main hardware environment configuration includes 
Intel Xeon processor, 64GB memory and Nvidia RTX2080Ti graphics card. All codes used in this manuscript have been open 
sourced in \url{https://github.com/Haodayu/Parallel-PINNs-with-Bidirectional-Balance}.

\subsection{Selection of Scaling Coefficients in BBM}
As mentioned in Section \ref{sec:backward balance module}, we propose BBM to make PINNs learning balanced. In actual applications, 
we first set up 50 repeated training experiments, all of which are trained for one epoch. The average loss values are recorder in 
Table \ref{Tab3}. It shows the rationality of our partition method.

\begin{table}[htbp]
  \caption{Initial value distribution for all classes}
  \label{Tab3}
  \vspace{-0.5cm}
  \begin{center}
      \resizebox{\linewidth}{!}{
      \begin{tabular}{|c|c|c|}
      \hline      
      \textbf{Class} & \textbf{$\mathcal{L}(C)$} & \textbf{range$(C)$} \\
      \hline
      {$C_1$} & {$\{ 0.78, 0.94, 0.62, 1.4, 1.2 \}\times 10^{6}$}
      & {$[6.24 \times 10^{5}, 1.42 \times 10^{6}]$} \\
      {$C_2$} & {$\{ 8.0, 8.6, 4.2, 0.11 \}\times 10^{9}$} 
      & {$[1.14 \times 10^{8}, 8.65 \times 10^{9}]$}\\
      {$C_3$} & {$\{ 1.3, 6.2, 0.063 \}\times 10^{16}$} 
      & {$[6.36 \times 10^{14}, 6.28 \times 10^{16}]$}\\
      \hline
      \end{tabular}}
  \end{center}
\end{table}

\par
\begin{figure}[htbp]
  \centering 
  \includegraphics[width=\linewidth]{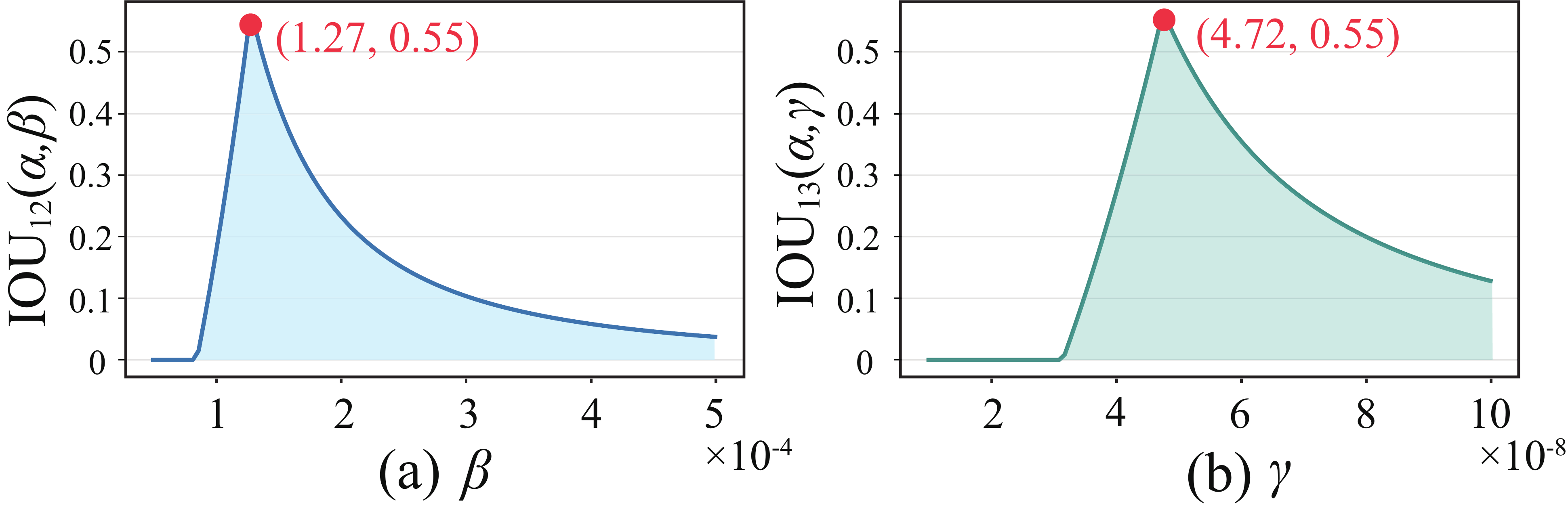}  
  \caption{Schematic of linear search results. The red dot indicates the optimal value.}
  \label{findscale}
  \end{figure}

If the scaling coefficient is too small, the loss will be scaled to near 0, which can also make IOU approach 1. At this time, 
various losses have reached some kind of balance, but too small loss value may cause vanishing gradient. This is obviously not 
what we want. To this end, we need to limit scaling coefficients. Based on Table \ref{Tab3}, we set $\alpha=1\times 10^{-2}$. 
The optimal value of other coefficients can be obtained by linear search (see Fig.~\ref{findscale}). When 
$\beta \approx 1.27 \times 10^{-4}$ and $\gamma \approx 4.72 \times 10^{-8}$, the value of IOU reaches maximum, and the three 
classes of loss almost achieve maximum balance.

\vspace{-0.5cm}
\begin{figure}[htbp]
  \centering  
  \includegraphics[width=0.45\textwidth]{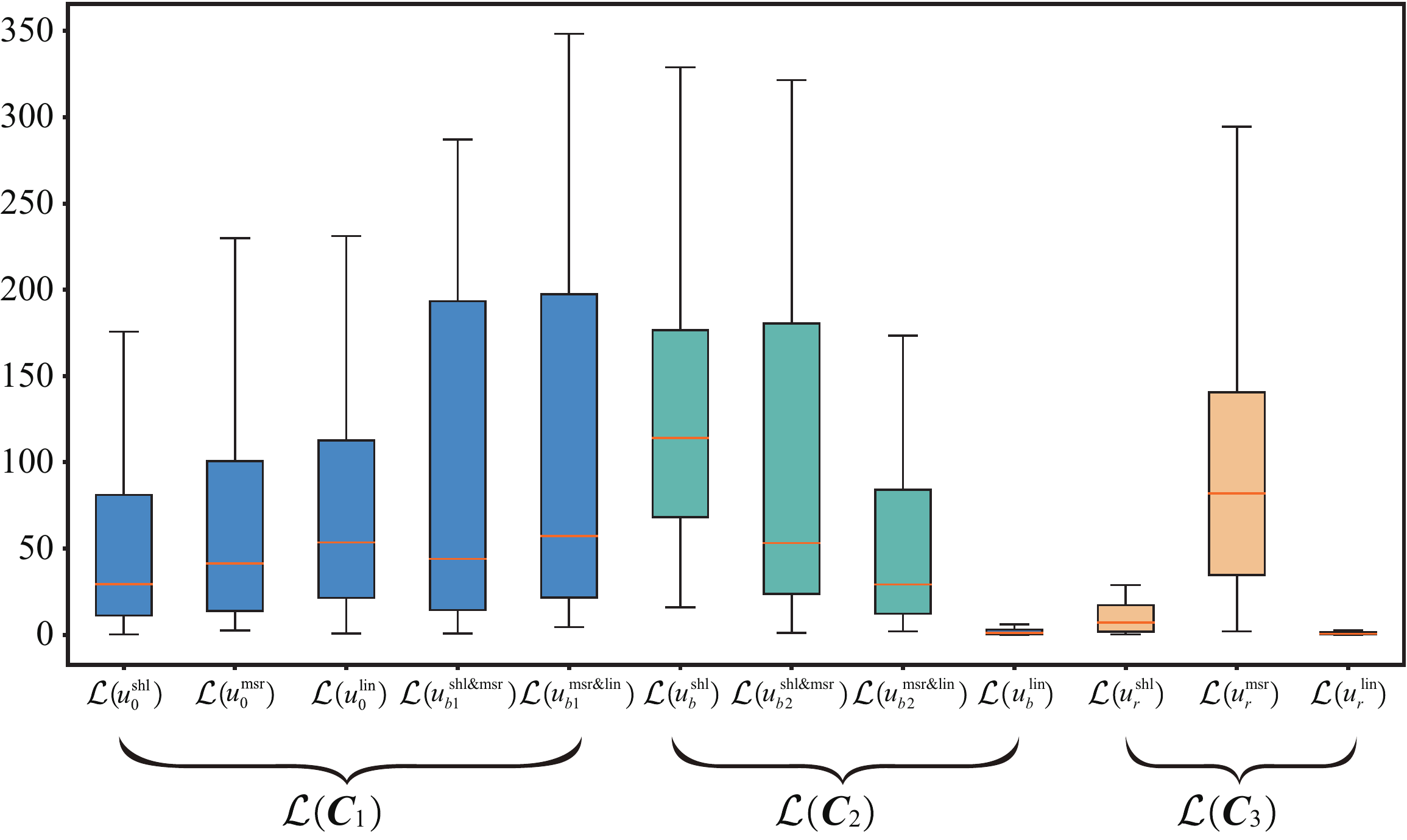}  
  \caption{Box plot of initial loss distribution after using BBM. The ordinate is the loss value under repeated experiments, and the abscissa is the loss item in different classes.}
  \label{findafter}
  \end{figure}

\begin{table*}[htbp]
  \caption{The result of MSE between ground truth and prediction under different model settings.}
  \label{Tab4}
  \begin{center}
      \begin{tabular}{|c|c|c|c|c|c|}
      \hline
      \multirow{2}{*}{\textbf{Model}} & \multirow{2}{*}{\textbf{Methods}} & \multicolumn{4}{|c|}{\textbf{Mean Square Error}} \\
      \cline{3-6}
      {} & {} & \textbf{SHL layer} & \textbf{MSR layer} & \textbf{LIN layer} & \textbf{Total} \\
      \hline
      {M1} & {PSF,FBM,BBM} & {      $5.2462 \times 10^{-5}$      } & {      $5.5566 \times 10^{-5}$      } & 
      {      $2.7532 \times 10^{-4}$      } & {      $1.9242 \times 10^{-4}$      } \\
      {M2} & {PSF,FBM} & {$1.0601 \times 10^{-1}$} & {$1.8373 \times 10^{-1}$} & {$1.6555 \times 10^{-1}$} & {$1.6022 \times 10^{-1}$} \\
      {M3} & {PSF,BBM} & {$6.7290 \times 10^{-1}$} & {$4.9186 \times 10^{-1}$} & {$2.4269 \times 10^{-1}$} & {$3.6442 \times 10^{-1}$} \\
      {M4} & {PSF} & {$6.9509 \times 10^{-1}$} & {$7.7504 \times 10^{-1}$} & {$3.5576 \times 10^{-1}$} & {$5.0050 \times 10^{-1}$} \\
      {M5} & {FBM,BBM} & {$4.0770 \times 10^{-2}$} & {$3.3856 \times 10^{-2}$} & {$7.1182 \times 10^{-2}$} & {$5.8265 \times 10^{-2}$} \\
      \hline
      \end{tabular}
  \end{center}
\end{table*}

\par
\begin{figure}[htbp]
  \centering  
  \includegraphics[width=\linewidth]{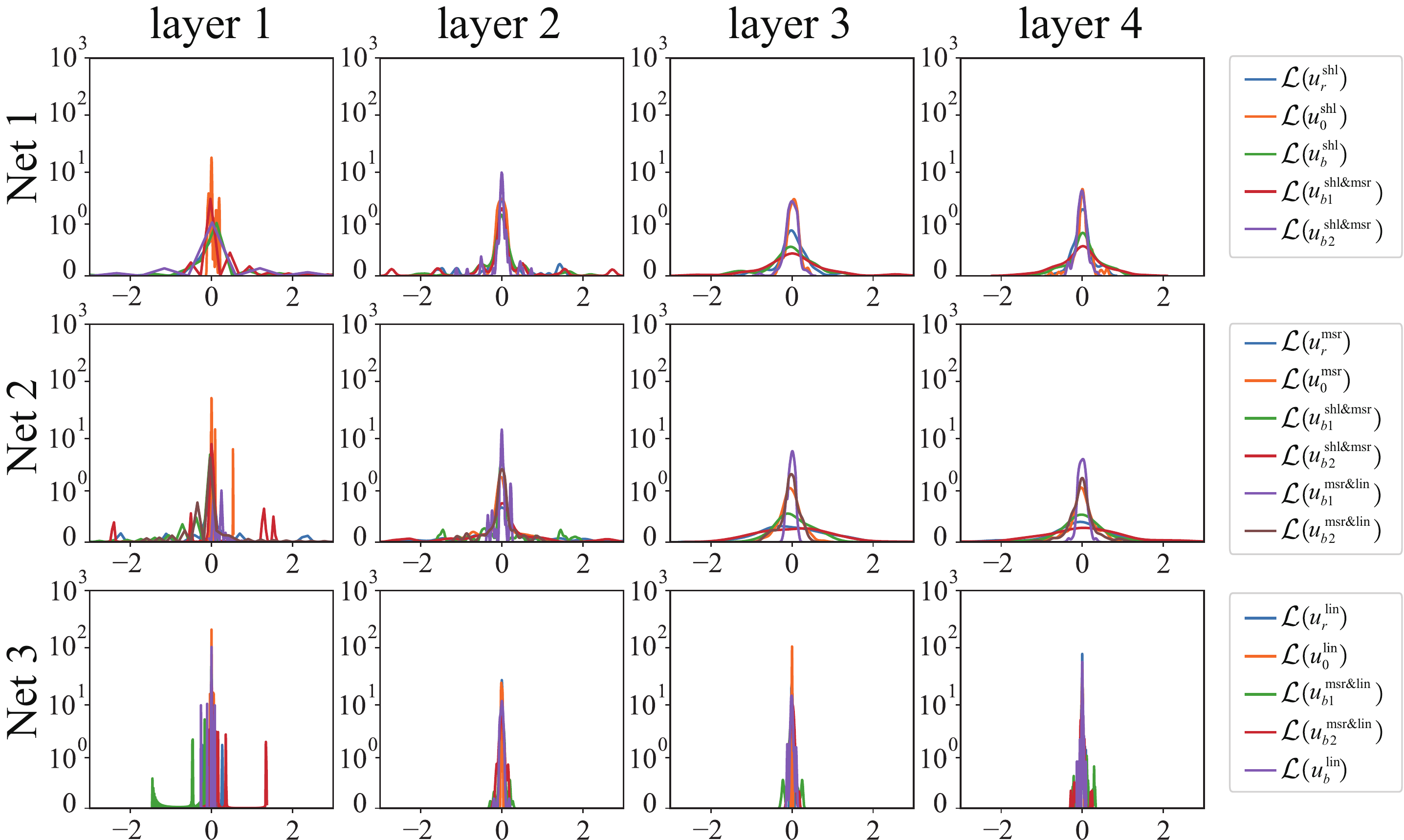}  
  \caption{Schematic of gradient flow after using BBM. The subgraph with m\textsuperscript{th} row and n\textsuperscript{th} column corresponds to the neuron gradient distribution curve of n\textsuperscript{th} layer in the m\textsuperscript{th} network. This is a snapshot at the 1000n\textsuperscript{th} ephocs.}
  \label{checkgrad}
  \end{figure}

To further visualize the impact of BBM, we track the gradient distribution of each loss during training. In Addition, we also 
track the gradient distribution of each loss during training. Fig.~\ref{findafter} and \ref{checkgrad} show that such a set of 
scaling coefficients effectively plays a balancing role, and enables the gradient flow of the training process to keep a good 
learning state.

\subsection{Ablation Experiments}
\begin{figure}[htbp]
  \centering
  \includegraphics[width=\linewidth]{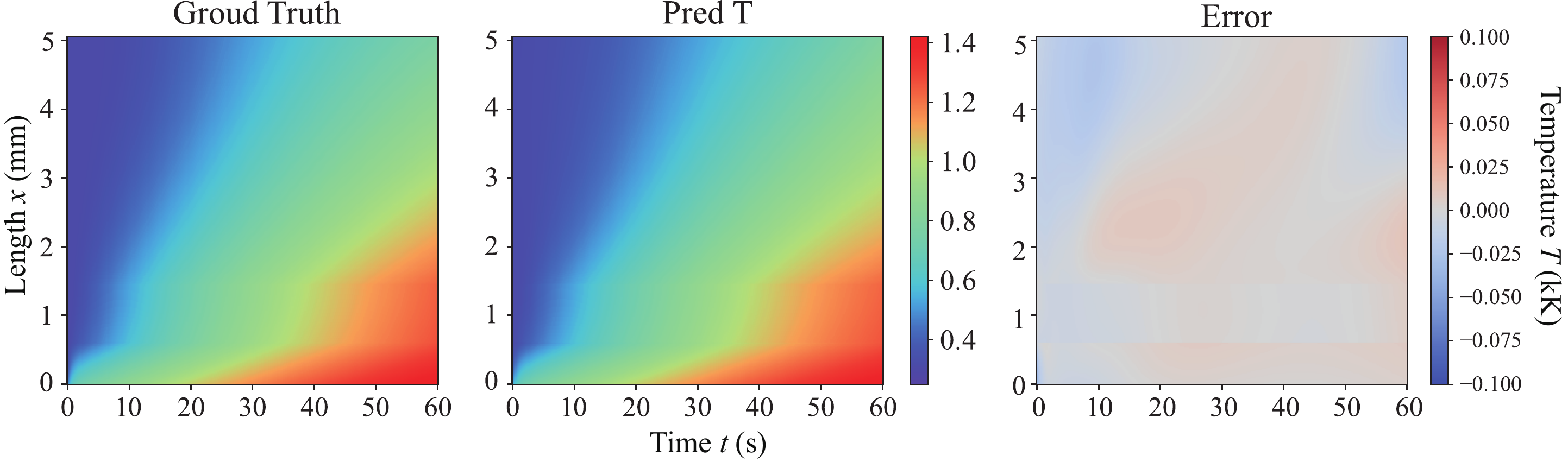}  
  \caption{2D visualization of M1 solution results. The temperature value uses color mapping and the unit is kilokelvin. The error diagram is the difference between the ground truth and the predicted value. }
  \label{vs60}
  \end{figure}

We designed a set of ablation experiments to verify the effectiveness of our methods. Each model was repeated three times randomly, 
all of which were trained for 20,000 epochs. The solving results are presented in Table \ref{Tab4}. Obviously, the M1 model with a 
complete framework achieves solving accuracies required for real-world. Fig.~\ref{vs60} shows that the changing trends of the 
ground truth and the predicted solution are almost the same.

\par
According to the poor performance of the M2-M4 models, we can know that the bidirectional balance is critical to successfully 
solving heat transfer problem of multilayer fabric. Using either FBM or BBM alone can slightly improve accuracy, but only when 
this two coexist can it produce a reliable and strong balance constraint in PINNs.
\begin{figure*}[htbp]
  \centering  
  \includegraphics[width=0.85\textwidth]{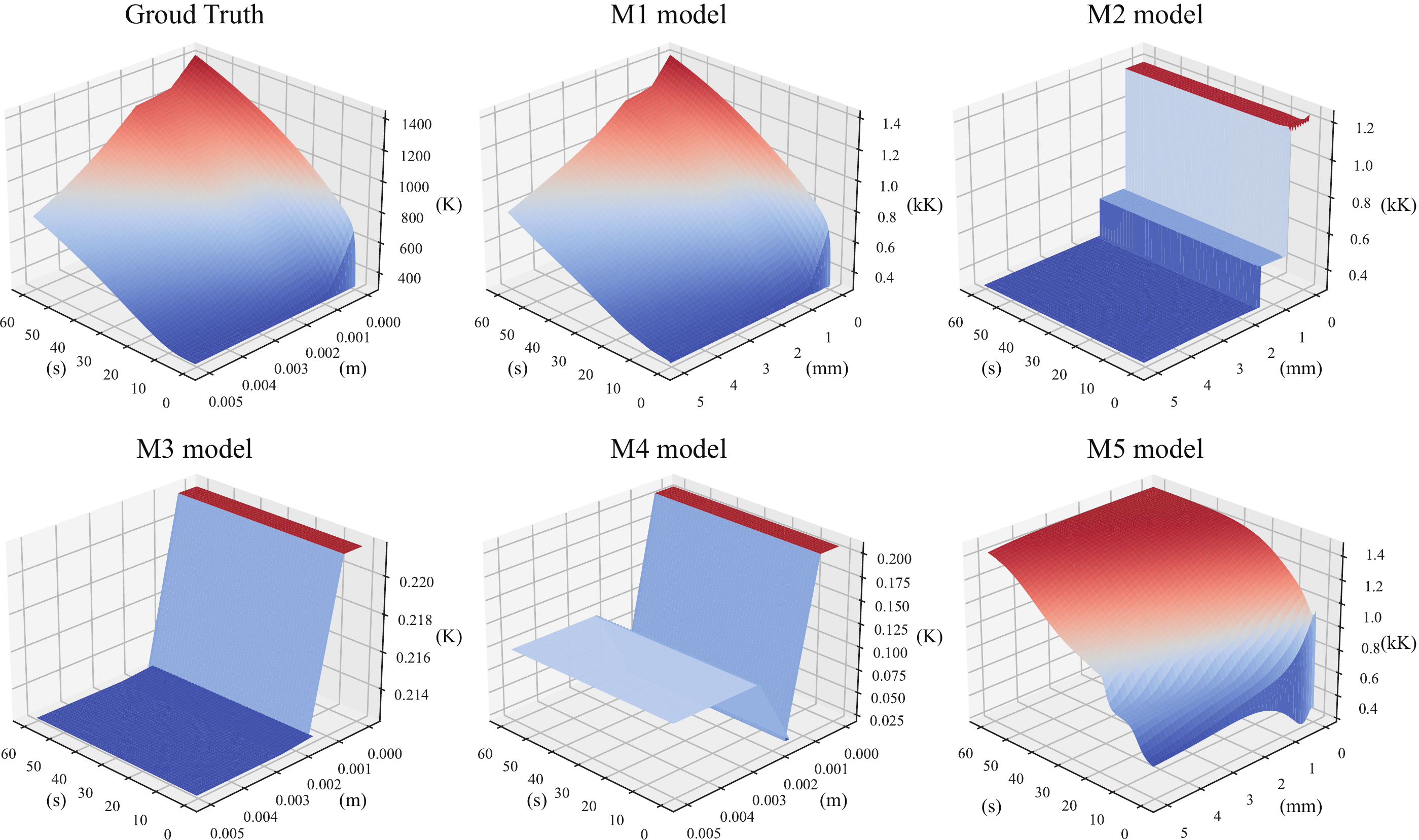}  
  \caption{3D visualization of M1-M5 models solution results. Whether the model uses FSM or not makes a difference in coordinate units.}
  \label{3dv}
  \end{figure*}

\par
Achieving M1 performance requires not only a suitable balance strategy, but also the high fitting capabilities of the parallel 
solving framework. Table \ref{Tab4} also shows the accuracy gap between the M5 and the M1 (use PSF or not). Fig.~\ref{3d} more intuitively 
reflects the superiority of using PSF in the coupled problem. As analyzed in Section \ref{sec:htm}, the temperature gradient 
between adjacent layers is not continuous and not constant. Therefore, the multiple networks and parallelism of PSF play a key 
role in fitting such sudden change. In conclusion, these experiments strongly proved the feasibility of our methods and emphasized 
that the maximum effect can only be achieved when these methods coexist.

\section{Summary and Outlook}
In this paper, we describe the problems of classical PINNs in a more general PDEs, and propose 
\emph{Parallel Physics-Informed Neural Networks with Bidirectional Balance} to solve it. Specifically, we focus on heat 
transfer problem in multilayer fabrics. On the one hand, this problem is composed of the coupled PDEs 
corresponding to multilayer fabrics, which brings a challenge to network fitting. On the other hand, 
the input and output physical quantities of the problem show a huge difference in magnitude, and the many 
conditions in the coupling system exacerbate this imbalance. Whether in the forward or backward process of 
the network, this imbalance will profoundly hinder training. For this reason, we propose a parallel solution 
framework to solve the difficulty of fitting coupled equations. Furthermore, two balancing strategies are 
proposed to be applied to the forward and backward processes respectively, so that the network can be trained 
correctly. The bidirectional balance method is proposed systematically for the first time. From the perspective 
of final performance, our approach is very successful in the heat transfer problem under different 
parameter settings.

\par
Despite there are enough theories to support our approach to other complex problems, we have not yet 
conducted empirical investigations on this. In addition, we use fully connected neural networks for 
all sub-networks, but we can also use other advanced network structures, such as convolutional neural 
networks or attention mechanism. These modifications can further improve the accuracy of the solution.

\appendices
\section{Expression of the Residual Terms}
\label{app:res}
According to the mathematical model of heat transfer of multilayer fabric, here we will give the expression of each residual term 
in the network, and thus derive the complex compound loss function in Section \ref{sec:parallel solving framework} of the paper. 
In addition, since the residual terms are affected by the forward balance module (FBM) mentioned in Section 
\ref{sec:forward balance module}, we will also give their transformed expression. First, we regard the physical base-units 
transformation as follows:
\allowdisplaybreaks
\begin{align*}
    T=&T'\times 10^{3}\\
    x=&x'\times 10^{-3}
\end{align*}
\par
Section \ref{sec:htm} of the paper analyzed the heat conduction and heat convection processes in the model. According 
to the existing physical equations, we can derive the residual terms and their transformation as follows:
\\
1) Initial conditions:
\begin{align*}
    u_{0}^{\text{shl}}:=&\left.T_{\text{shl}}\right|_{t=0}-T_{0}\\
    =&(\left.T'_{\text{shl}}\right|_{t=0}-T'_{0})\times 10^{3}\\
    u_{0}^{\text{msr}}:=&\left.T_{\text{msr}}\right|_{t=0}-T_{0} \\
    =&(\left.T'_{\text{msr}}\right|_{t=0}-T'_{0})\times 10^{3}\\
    u_{0}^{\text {lin }}:=&\left.T_{\text{lin}}\right|_{t=0}-T_{0} \\
    =&(\left.T'_{\text{lin}}\right|_{t=0}-T'_{0})\times 10^{3}
\end{align*}
2) Thermal diffusion control equations:
\begin{align*}
    u_{r}^{\text{shl}}:=&C_{\text{shl}}^{\text{A}} \frac{\partial T_{\text {shl }}}{\partial t}-k_{\text {shl}} \frac{\partial^{2} T_{\text {shl }}}{\partial x^{2}}\\
    =&C_{\text{shl}}^{\text{A}} \frac{\partial T'_{\text{shl}}}{\partial t} \times 10^{3}-k_{\text{shl}} \frac{\partial^{2} T'_{\text{shl}}}{\partial x'^2} \times 10^{9}\\
    u_{r}^{\text{msr}}:=&C_{\text{msr}}^{\text{A}} \frac{\partial T_{\text{msr}}}{\partial t}-k_{\text{msr}} \frac{\partial^{2} T_{\text{msr}}}{\partial x^{2}} \\
    =&C_{\text{msr}}^{\text{A}} \frac{\partial T'_{\text{msr}}}{\partial t} \times 10^{3} -k_{\text{msr}} \frac{\partial^{2} T'_{\text{msr}}}{\partial x'^{2}} \times 10^{9}\\
    u_{r}^{\text{lin}}:=&C_{\text{lin}}^{\text{A}} \frac{\partial T_{\text{lin}}}{\partial t}-k_{\text{lin}} \frac{\partial^{2} T_{\text {lin}}}{\partial x^{2}} \\
    =&C_{\text{lin}}^{\text{A}} \frac{\partial T'_{\text{lin}}}{\partial t} \times 10^{3} -k_{\text{lin}} \frac{\partial^{2} T'_{\text {lin}}}{\partial x'^{2}} \times 10^{9}
\end{align*}
3) Boundary conditions of outer surfaces: 
\begin{align*}
    u_{b}^{\text{shl}}:=&-\left.k_{\text{shl}} \frac{\partial T_{\text{shl}}}{\partial x}\right|_{x=0}-h_{\text{g}}\left(T_{\text{g}}-\left.T_{\text{shl}}\right|_{x=0}\right)\\
    =&-\left.k_{\text{shl}}\frac{\partial T'_{\text{shl}}}{\partial x'}\right|_{x'=0}\times 10^{6}\\
    &-h_{\text{g}}\left(T'_{\text{g}}-\left.T'_{\text{shl}}\right|_{x'=0}\right) \times 10^{3}\\
    u_{b}^{\text {lin }}:=&-\left.k_{\text{lin}} \frac{\partial T_{\text{lin}}}{\partial x}\right|_{x=L_{\text{fab}}}-h_{\text{air}}\left(\left.T_{\text{lin}}\right|_{x=L_{\text{fab}}}-T_{0}\right)\\
    =&-\left.k_{\text{lin}}\frac{\partial T'_{\text{lin}}}{\partial x'}\right|_{x'=L'_{\text{fab}}} \times 10^{6}\\
    &-h_\text{air}\left(\left.T'_{\text{lin}}\right|_{x'=L'_{\text{fab}}}-T'_{0}\right) \times 10^{3}
\end{align*}
4) Boundary conditions of adjacent surfaces: 
\begin{align*}
    u_{b1}^{\text{shl\&msr}}:=&\left.T_{\text {shl}}\right|_{x=L_{\text {shl}}}-\left.T_{\text{msr}}\right|_{x=L_{\text {shl}}}\\
    =&(\left.T'_{\text{shl}}\right|_{x'=L'_{\text{shl}}}-\left.T'_{\text{msr}}\right|_{x'=L'_{\text{shl}}}) \times 10^{3}\\
    u_{b1}^{\text {msr\&lin}}:=&\left.T_{\text{msr}}\right|_{x=L_{\text{msr}}}-\left.T_{\text{lin}}\right|_{x=L_{\text{msr}}} \\
    =&(\left.T'_{\text{msr}}\right|_{x'=L'_{\text{msr}}}-\left.T'_{\text{lin}}\right|_{x'=L'_{\text{msr}}}) \times 10^{3}\\
    u_{b2}^{\text{shl\&msr}}:=&\left.k_{\text {shl}} \frac{\partial T_{\text{shl}}}{\partial x}\right|_{x=L_{\text{shl}}}-\left.k_{\text{msr}} \frac{\partial T_{\text{msr}}}{\partial x}\right|_{x=L_{\text{shl}}} \\
    =&(\left.k_{\text{shl}}\frac{\partial T'_{\text{shl}}}{\partial x'}\right|_{x'=L'_{\text{shl}}}-\left.k_{\text{msr}} \frac{\partial T'_{\text{msr}}}{\partial x'}\right|_{x'=L'_{\text{shl}}}) \times 10^{6}\\
    u_{b2}^{\text {msr\&lin}}:=&\left.k_{\text{msr}} \frac{\partial T_{\text{msr}}}{\partial x}\right|_{x=L_{\text{msr}}}-\left.k_{\text{lin}} \frac{\partial T_{\text{lin}}}{\partial x}\right|_{x=L_{\text{msr}}} \\
    =&(\left.k_{\text{msr}} \frac{\partial T'_{\text{msr}}}{\partial x}\right|_{x'=L'_{\text{msr}}}-\left.k_{\text{lin}} \frac{\partial T'_{\text{lin}}}{\partial x'}\right|_{x'=L'_{\text{msr}}}) \times 10^{6}
\end{align*}

\section{Additional Experiments}
In extreme flash fire environments, even thermal protective clothing can only be effective for the first 
tens of seconds. During this time, the temperature changes drastically, which is a challenge to our proposed 
numerical solving method. To verify this performance, we add experiments with various time settings. Other 
settings are the same as M1 in the original text. The solving accuracy and 2D visualization results are shown below.
\begin{table}[htbp]
  \caption{Mean square error between ground truth and prediction under different model settings.}
  \label{Tab5}
  \vspace{-0.7cm}
  \begin{center}
    \resizebox{\linewidth}{!}{\begin{tabular}{|c|c|c|c|c|}  
      \hline
      {\textbf{Time}} &{10s} &{30s} &{60s} &{120s} \\
      \hline
      {\textbf{SHL}} & {$5.2382 \times 10^{-6}$} & {$1.5055 \times 10^{-5}$} & {$5.8359 \times 10^{-5}$} & {$3.1514 \times 10^{-4}$} \\ 
      {\textbf{MSR}} & {$2.3912 \times 10^{-6}$} & {$5.1216 \times 10^{-6}$} & {$8.2250 \times 10^{-5}$} & {$3.7955 \times 10^{-4}$} \\  
      {\textbf{LIN}} & {$1.8996 \times 10^{-5}$} & {$1.2241 \times 10^{-4}$} & {$1.9701 \times 10^{-4}$} & {$1.7403 \times 10^{-3}$} \\
      {\textbf{Total}} & {$1.3214 \times 10^{-5}$} & {$7.9984 \times 10^{-5}$} & {$1.5024 \times 10^{-4}$} & {$1.2199 \times 10^{-3}$} \\
      \hline
      \end{tabular}}
  \end{center}
\end{table}
\begin{figure}[htb]
  \centering  
  \includegraphics[width=0.9\linewidth]{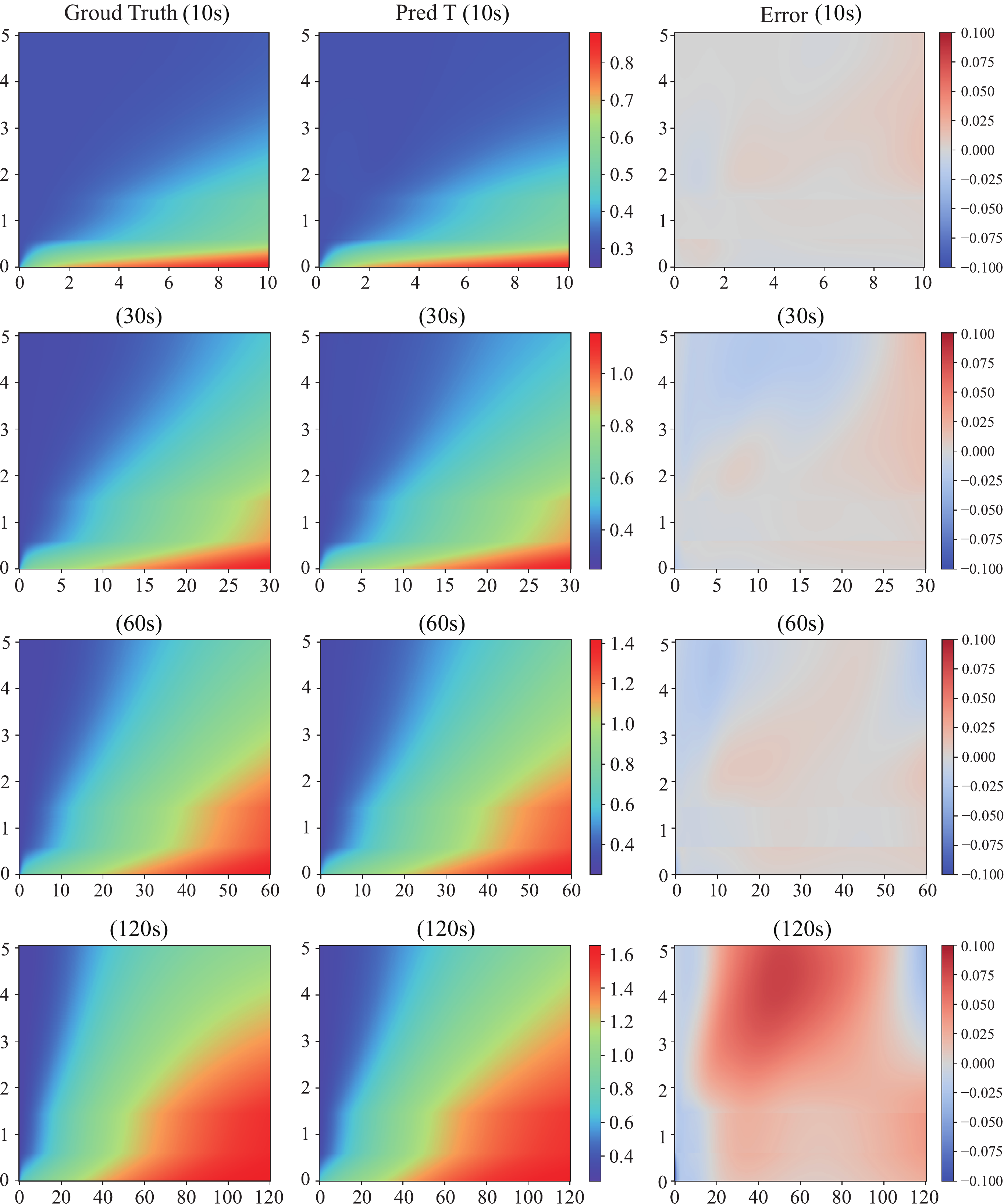} 
  \caption{2D Visualization of M1 solving results with different time settings.}
  \label{vs}
\end{figure}

\bibliographystyle{IEEEtran}
\bibliography{IEEEabrv, PINNref}

\begin{thebibliography}{10}
\providecommand{\url}[1]{#1}
\csname url@samestyle\endcsname
\providecommand{\newblock}{\relax}
\providecommand{\bibinfo}[2]{#2}
\providecommand{\BIBentrySTDinterwordspacing}{\spaceskip=0pt\relax}
\providecommand{\BIBentryALTinterwordstretchfactor}{4}
\providecommand{\BIBentryALTinterwordspacing}{\spaceskip=\fontdimen2\font plus
\BIBentryALTinterwordstretchfactor\fontdimen3\font minus
  \fontdimen4\font\relax}
\providecommand{\BIBforeignlanguage}[2]{{%
\expandafter\ifx\csname l@#1\endcsname\relax
\typeout{** WARNING: IEEEtran.bst: No hyphenation pattern has been}%
\typeout{** loaded for the language `#1'. Using the pattern for}%
\typeout{** the default language instead.}%
\else
\language=\csname l@#1\endcsname
\fi
#2}}
\providecommand{\BIBdecl}{\relax}
\BIBdecl

\bibitem{2012ImageNet}
A.~Krizhevsky, I.~Sutskever, and G.~Hinton, ``Imagenet classification with deep
  convolutional neural networks,'' \emph{Advances in neural information
  processing systems}, vol.~25, no.~2, 2012.

\bibitem{2014Learning}
K.~Cho, B.~V. Merrienboer, C.~Gulcehre, D.~BaHdanau, F.~Bougares, H.~Schwenk,
  and Y.~Bengio, ``Learning phrase representations using rnn encoder-decoder
  for statistical machine translation,'' \emph{Computer Science}, 2014.

\bibitem{2015Predicting}
B.~Alipanahi, A.~Delong, M.~T. Weirauch, and B.~J. Frey, ``Predicting the
  sequence specificities of dna- and rna-binding proteins by deep learning,''
  \emph{Nature Biotechnology}, 2015.

\bibitem{1996Evaluating}
K.~Jambunathan, S.~L. Hartle, S.~Ashforth-Frost, and V.~N. Fontama,
  ``Evaluating convective heat transfer coefficients using neural networks,''
  \emph{International Journal of Heat Mass Transfer}, vol.~39, no.~11, pp.
  2329--2332, 1996.

\bibitem{owhadi2015bayesian}
H.~{Owhadi}, ``Bayesian numerical homogenization,'' \emph{Multiscale Modeling
  and Simulation}, vol.~13, no.~3, pp. 812--828, 2015.

\bibitem{fonda2019deep}
E.~{Fonda}, A.~{Pandey}, J.~{Schumacher}, and K.~R. {Sreenivasan}, ``Deep
  learning in turbulent convection networks.'' \emph{Proceedings of the
  National Academy of Sciences of the United States of America}, vol. 116,
  no.~18, pp. 8667--8672, 2019.

\bibitem{raissi2019physics}
M.~{Raissi}, P.~{Perdikaris}, and G.~E. {Karniadakis}, ``Physics-informed
  neural networks: A deep learning framework for solving forward and inverse
  problems involving nonlinear partial differential equations,'' \emph{Journal
  of Computational Physics}, vol. 378, pp. 686--707, 2019.

\bibitem{raissi2020hidden}
M.~{Raissi}, A.~{Yazdani}, and G.~E. {Karniadakis}, ``Hidden fluid mechanics:
  Learning velocity and pressure fields from flow visualizations.''
  \emph{Science}, vol. 367, no. 6481, pp. 1026--1030, 2020.

\bibitem{sun2020surrogate}
L.~{Sun}, H.~{Gao}, S.~{Pan}, and J.-X. {Wang}, ``Surrogate modeling for fluid
  flows based on physics-constrained deep learning without simulation data,''
  \emph{Computer Methods in Applied Mechanics and Engineering}, vol. 361, p.
  112732, 2020.

\bibitem{raissi2019deep}
M.~{Raissi}, Z.~{Wang}, M.~S. {Triantafyllou}, and G.~E. {Karniadakis}, ``Deep
  learning of vortex-induced vibrations,'' \emph{Journal of Fluid Mechanics},
  vol. 861, pp. 119--137, 2019.

\bibitem{costabal2020physics}
F.~S. {Costabal}, Y.~{Yang}, P.~{Perdikaris}, D.~E. {Hurtado}, and E.~{Kuhl},
  ``Physics-informed neural networks for cardiac activation mapping,''
  \emph{Frontiers of Physics in China}, vol.~8, p.~42, 2020.

\bibitem{kissas2020machine}
G.~{Kissas}, Y.~{Yang}, E.~{Hwuang}, W.~R. {Witschey}, J.~A. {Detre}, and
  P.~{Perdikaris}, ``Machine learning in cardiovascular flows modeling:
  Predicting arterial blood pressure from non-invasive 4d flow mri data using
  physics-informed neural networks,'' \emph{Computer Methods in Applied
  Mechanics and Engineering}, vol. 358, p. 112623, 2020.

\bibitem{sirignano2018dgm}
J.~{Sirignano} and K.~{Spiliopoulos}, ``Dgm: A deep learning algorithm for
  solving partial differential equations,'' \emph{Journal of Computational
  Physics}, vol. 375, pp. 1339--1364, 2018.

\bibitem{han2018solving}
J.~{Han}, A.~{Jentzen}, and W.~{E}, ``Solving high-dimensional partial
  differential equations using deep learning,'' \emph{Proceedings of the
  National Academy of Sciences of the United States of America}, vol. 115,
  no.~34, pp. 8505--8510, 2018.

\bibitem{fuks2020limitations}
O.~{Fuks} and H.~A. {Tchelepi}, ``Limitations of physics informed machine
  learning for nonlinear two-phase transport in porous media,'' \emph{Journal
  of Machine Learning for Modeling and Computing}, vol.~1, no.~1, pp. 19--37,
  2020.

\bibitem{wang2020understanding}
S.~{Wang}, Y.~{Teng}, and P.~{Perdikaris}, ``Understanding and mitigating
  gradient pathologies in physics-informed neural networks.'' \emph{arXiv
  preprint arXiv:2001.04536}, 2020.

\bibitem{udayraj2016heat}
{Udayraj}, P.~{Talukdar}, A.~{Das}, and R.~{Alagirusamy}, ``Heat and mass
  transfer through thermal protective clothing – a review,''
  \emph{International Journal of Thermal Sciences}, vol. 106, pp. 32--56, 2016.

\bibitem{torvi1999heat}
D.~A. {Torvi} and J.~D. {Dale}, ``Heat transfer in thin fibrous materials under
  high heat flux,'' \emph{Fire Technology}, vol.~35, no.~3, pp. 210--231, 1999.

\bibitem{chitrphiromsri2003modeling}
P.~{Chitrphiromsri} and A.~V. {Kuznetsov}, ``Modeling heat and moisture
  transport in firefighter protective clothing during flash fire exposure,''
  \emph{Heat and Mass Transfer}, vol.~41, no.~3, pp. 206--215, 2003.

\bibitem{zhu2009modeling}
F.-L. Zhu and W.-Y. Zhang, ``Modeling heat transfer for heat-resistant fabrics
  considering pyrolysis effect under an external heat flux,'' \emph{Journal of
  Fire Sciences}, vol.~27, no.~1, pp. 81--96, 2009.

\bibitem{ghazy2012numerical}
A.~{Ghazy} and D.~J. {Bergstrom}, ``Numerical simulation of heat transfer in
  firefighters' protective clothing with multiple air gaps during flash fire
  exposure,'' \emph{Numerical Heat Transfer Part A-applications}, vol.~61,
  no.~8, pp. 569--593, 2012.

\bibitem{paszke2019pytorch}
A.~{Paszke}, S.~{Gross}, F.~{Massa}, A.~{Lerer}, J.~{Bradbury}, and et~al.,
  ``Pytorch: An imperative style, high-performance deep learning library,'' in
  \emph{Advances in Neural Information Processing Systems}, vol.~32, 2019, pp.
  8026--8037.

\bibitem{1995THE}
Y.~Han, ``The existence and uniqueness of the global solutions for the mixed
  problem of some nonlinear heat equations,'' \emph{Journal of Sichuan Normal
  University (NATURAL SCIENCE)}, 1995.

\bibitem{hornik1989multilayer}
K.~{Hornik}, M.~{Stinchcombe}, and H.~{White}, ``Multilayer feedforward
  networks are universal approximators,'' \emph{Neural Networks}, vol.~2,
  no.~5, pp. 359--366, 1989.

\bibitem{siegel2020approximation}
J.~W. {Siegel} and J.~{Xu}, ``Approximation rates for neural networks with
  general activation functions,'' \emph{Neural Networks}, vol. 128, pp.
  313--321, 2020.

\bibitem{evans2010partial}
L.~C. Evans, \emph{Partial Differential Equations}.\hskip 1em plus 0.5em minus
  0.4em\relax American Mathematical Society, 2010.

\bibitem{kingma2015adam}
D.~P. {Kingma} and J.~L. {Ba}, ``Adam: A method for stochastic optimization,''
  in \emph{ICLR 2015 : International Conference on Learning Representations
  2015}, 2015.

\bibitem{he2015delving}
K.~{He}, X.~{Zhang}, S.~{Ren}, and J.~{Sun}, ``Delving deep into rectifiers:
  Surpassing human-level performance on imagenet classification,'' in
  \emph{2015 IEEE International Conference on Computer Vision (ICCV)}, 2015,
  pp. 1026--1034.

\end{thebibliography}

\end{document}